\documentclass[letterpaper, 10 pt, journal]{ieeeconf}

\IEEEoverridecommandlockouts                              % This command is only needed if 
\usepackage{etoolbox}
\makeatletter
\patchcmd{\@makecaption}
  {\scshape}
  {}
  {}
  {}
\makeatletter
\patchcmd{\@makecaption}
  {\\}
  {.\ }
  {}
  {}
\makeatother

\usepackage{graphics} % for pdf, bitmapped graphics files
\usepackage{epsfig} % for postscript graphics files
\usepackage{mathptmx} % assumes new font selection scheme installed
\usepackage{times} % assumes new font selection scheme installed
\usepackage{amsmath} % assumes amsmath package installed
\usepackage{amssymb}  % assumes amsmath package installed
\usepackage{algorithm}
\usepackage{algpseudocode}
\usepackage{cite}
\usepackage{hyperref}
\usepackage{booktabs}
\usepackage{caption}
\usepackage{makecell} 
\usepackage{multirow}
\usepackage{adjustbox}
\usepackage{changepage}
\usepackage{xcolor}
\usepackage{wrapfig}
\usepackage{subcaption}

\usepackage{url}
\usepackage{graphicx}
\usepackage{siunitx}
\usepackage{bm}
\usepackage{mathtools}
\usepackage{amsfonts}
\usepackage{wrapfig}

\usepackage{listings}

\let\labelindent\relax

\usepackage{enumitem}

\usepackage{geometry}
\graphicspath{{fig/}}

\hypersetup{
  colorlinks   = true,    % Colours links instead of ugly boxes
  urlcolor     = blue,    % Colour for external hyperlinks
  linkcolor    = blue,    % Colour of internal links
  citecolor    = blue      % Colour of citations
}

\setlist[itemize]{leftmargin=*}

\title{\LARGE \bf Latent Embedding Adaptation for Human Preference Alignment in Diffusion Planners}

\author{Wen Zheng Terence Ng$^{1,2}$, Jianda Chen$^{1}$, Yuan Xu$^{1}$, Tianwei Zhang$^{1}$ % <-this % stops a space
\\
{\small$^1$Nanyang Technological University, $^2$Continental Automotive Singapore}
\\
{\tt\small\{ngwe0099, jianda001, xu.yuan, tianwei.zhang\}@ntu.edu.sg }\\ 
}

\begin{document}

\maketitle
\thispagestyle{empty}
\pagestyle{empty}

%%%%%%%%%%%%%%%%%%%%%%%%%%%%%%%%%%%%%%%%%%%%%%%%%%%%%%%%%%%%%%%%%%%%%%%%%%%%%%%%
\begin{abstract}

This work addresses the challenge of personalizing trajectories generated in automated decision-making systems by introducing a resource-efficient approach that enables rapid adaptation to individual users' preferences. 
Our method leverages a pretrained conditional diffusion model with Preference Latent Embeddings (PLE), trained on a large, reward-free offline dataset. The PLE serves as a compact representation for capturing specific user preferences. By adapting the pretrained model using our proposed preference inversion method, which directly optimizes the learnable PLE, we achieve superior alignment with human preferences compared to existing solutions like Reinforcement Learning from Human Feedback (RLHF) and Low-Rank Adaptation (LoRA). To better reflect practical applications, we create a benchmark experiment using real human preferences on diverse, high-reward trajectories.
\end{abstract}

% \keywords{Preference Learning, Diffusion, Offline Reinforcement Learning} 

%===============================================================================

\section{Introduction}

In today's increasingly automated world, personalization is crucial for decision-making systems to effectively cater to individual's needs, preferences, and circumstances. Tailoring experiences enhances the system effectiveness and user satisfaction across diverse applications, such as customizing self-driving vehicles \cite{bae2020self,he2023toward}, transforming robotic assistants into adaptive companions \cite{ohnbar2018personalized,gao2018robot,moro2018learning,wen2019online,woodworth2018preference}, and optimizing prosthetics for wearers' unique requirements \cite{tu2021data, zhang2022imposing, nalam2022admittance}. However, accurately capturing and aligning the abstract and dynamic human preferences with automated systems remains a complex challenge \cite{wirth2017survey,christiano2017deep, wilson2012bayesian}.

\begin{figure}[t]
    \centering
    \includegraphics[width=0.7\linewidth]{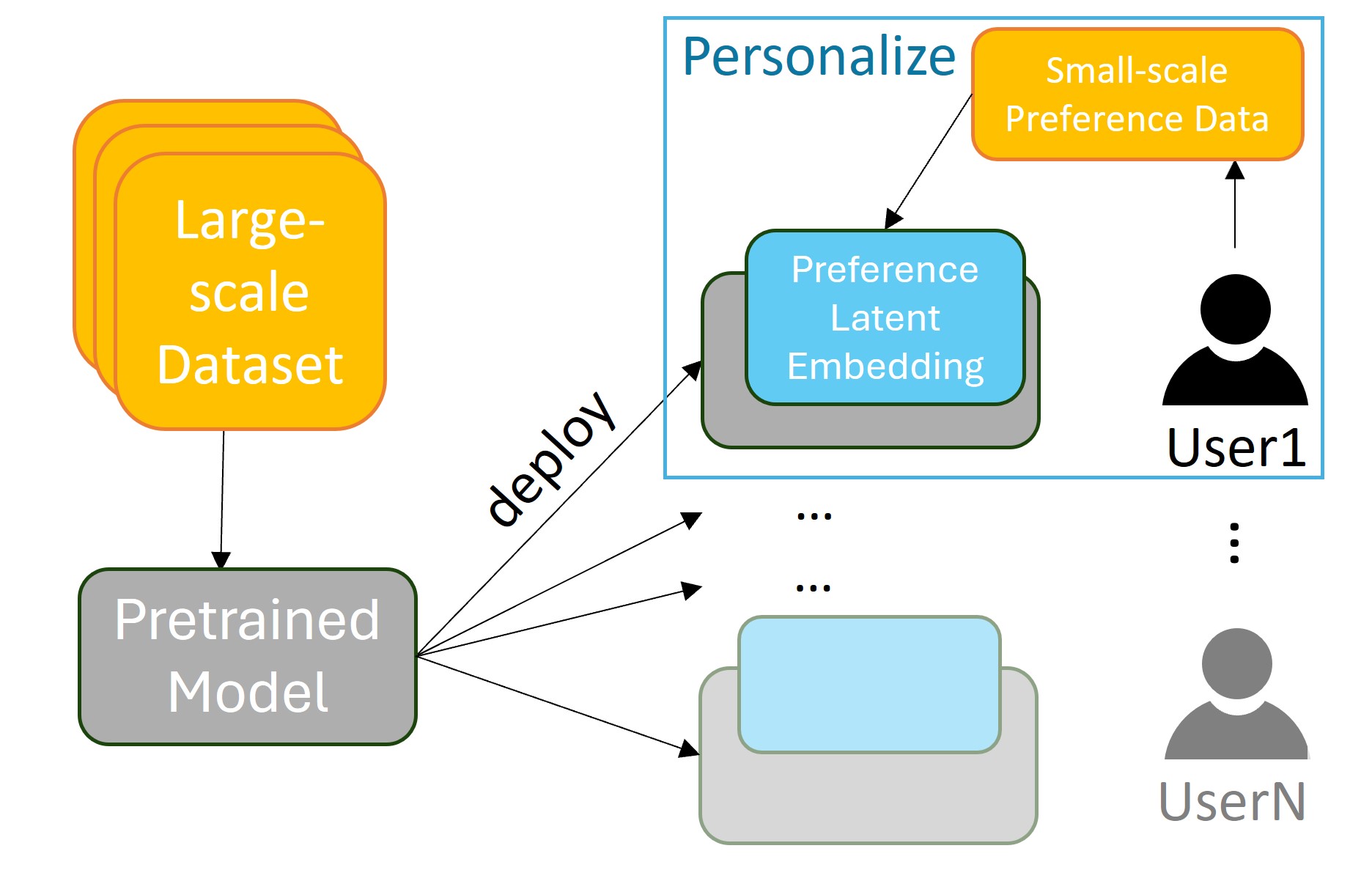}
    \caption{\textbf{Overview of personalizing decision-making models.} We leverage large-scale offline data for pretraining, followed by rapid and efficient personalization using small-scale preference data.}
    \label{fig:overview}
\vspace{-10pt}
\end{figure}

This work tackles this personalization challenge in trajectories generated by automated decision-making systems, aiming to create adaptable and reusable models that cater to individual user's needs \cite{haydari2020deep,yu2021reinforcement,zhang2019deep,nian2020review}. While large-scale pretrained models offer broad capabilities \cite{touvron2023llama,brown2020language,reed2022generalist,dosovitskiy2020image}, they lack the individual customization, and training personalized models for every user is infeasible. In contrast, it is more promising to first pretrain a model on large-scale offline data, and then align it with human preferences using smaller, user-specific preference datasets, as shown in Figure \ref{fig:overview}. The adoption of pretrained models from offline data avoids costly or risky direct interaction, enabling broader applications in challenging environments \cite{zareleveraging, fan2024learn, codevilla2018end}.
The process for adapting human preferences must be computationally efficient, enabling updates for many users and deployment on edge devices, through minimal data requirements.
Therefore, we adopt this pretrain-align framework to achieve personalized decision making.

However, there are still a couple of difficulties to realize this system. (1) For pretraining the decision-making models, some approaches~\cite{lee2021pebble, eysenbach2018diversity} perform the training without rewards, but require the online interaction with the environment, which is not applicable in our offline setting. Other approaches with offline reinforcement learning (RL)~\cite{levine2020offline,kumar2020conservative, kostrikov2022offline, fujimoto2021minimalist} relies on rewards, which are often unavailable or difficult to quantify for human preferences. It is important but challenging to address both requirements simultaneously. (2) For adapting models to human preferences, RLHF ~\cite{christiano2017deep} has emerged as a key technique for integrating human preferences into decision-making systems~\cite{knox2022models, hejna2023few, ibarz2018reward, kim2023preference}.
It works by first learning reward models to capture individual preferences and then refining policies based on those learned reward models. 
\emph{Direct policy optimization} (DPO) offers an alternative approach by directly aligning policies with human preferences, bypassing the need for a separate reward model \cite{rafailov2023direct}. However, both RLHF and DPO face computational challenges due to the large number of parameters involved during alignment, making them less resource-efficient. Furthermore, both methods require careful tuning to prevent the adapted model from deviating too far from the base model.

We propose a pretrain-align framework to enable efficient and rapid personalized decision-making. Our solution is built atop of diffusion-based planners~\cite{janner2022diffuser}, which leverage the expressive power of diffusion models~\cite{ho2020denoising} to learn flexible and tractable models for trajectory generation. 
We introduce \emph{preference latent embeddings} (PLE), low-dimensional vectors that effectively encode human preferences, for rapidly adapting pretrained models to individual user preferences. Our method involves three stages: (1) Pretraining a diffusion model without reward supervision using a large state-action-only sequence dataset, (2) adapting the model to specific preferences using a small set of human labels, and (3) generating trajectories aligned with the learned preferences.
Results demonstrate that our method adapts more accurately to human preferences with less data, both in offline datasets and in our custom dataset with real human labels.
% (rewrite) exp + contri
Our key contributions can be summarized as follows:
\begin{itemize} %[leftmargin=*,noitemsep,topsep=0pt,parsep=0pt,partopsep=0pt]
    \item Introducing a reward-free pretraining approach that jointly learns meaningful representations for PLE.
    \item Proposing an adaptation method for rapid preference alignment through preference inversion.
    \item Creating a benchmark experiment using real human preferences on diverse, high-reward trajectories.
    \item Conducting detailed evaluations and ablation studies using both our human-annotated dataset and an existing dataset.

\end{itemize}

%===============================================================================

\section{Background and Related Work}
\label{sec:background}

%================================================================================

\noindent\textbf{Diffusion Probabilistic Models (DPMs).}
These powerful and versatile generative models offer a high degree of flexibility and tractability in modeling complex data distributions \cite{sohl2015deep, ho2020denoising}. 
The core principle behind DPMs is to learn to reverse a diffusion process by progressively denoising data points that have been transformed into random noise through a forward Markov chain. Given a data point sampled from a real data distribution, $\mathbf{x}_0 \sim q(\mathbf{x})$, this chain is defined by $q(x_k | x_{k-1}) = \mathcal{N}(x_k | \sqrt{\alpha_k} x_{k-1}, (1-\alpha_k)I)$, where $\mathcal{N}(\mu, \Sigma)$ represents a Gaussian distribution with mean $\mu$ and covariance $\Sigma$, and $\alpha_k$ determines the noise schedule with discrete noise time-step $k$.
From these noise-augmented data points, a variational reverse Markov chain, parameterized by \( p_\theta(x_{k-1} | x_k) = \mathcal{N}(x_{k-1} | \mu_\theta(x_k, k), \Sigma_\theta(x_k, k)) \), is used to reconstruct the original data point \( x_0 \).
\cite{ho2020denoising} introduced an optimized surrogate loss function to simply this process:
\begin{equation}
\label{eqn:ddpm}
\mathcal{L}(\theta)=\mathbb{E}_{k \sim[1, K], x_0 \sim q, {\epsilon} \sim \mathcal{N}(0, I)}\left[\left\|{\epsilon}-{\epsilon}_\theta\left(x_k, k\right)\right\|^2\right],
\end{equation}
where ${\epsilon}$ is the sampled noise and ${\epsilon}_\theta$ is the noise predicting model. 
By optimizing this loss, new data points can be generated through a sampling process via the forward Markov chain with $\mu_\theta(x_k, k) = \frac{1}{\sqrt{\alpha_k}}\left(x_k - \frac{1-\alpha_k}{\sqrt{1-\bar{\alpha}k}} {\epsilon}_\theta(x_k, k)\right)$.

%\noindent\textbf{Conditional DPMs}.
DPMs can be extended to conditional generation using the \emph{classifier-free} guidance \cite{Ho2022ClassifierFreeDG}, which enables the conditional model, $p_\theta(x_{k-1}|x_k, c)$, to generate samples conditioned on a context input $c$.
During sampling, the predicted noise is adapted to a weighted combination of conditional and non-conditional sampling:
% \begin{equation}
% \label{eqn:ddpm_sample}
$
\hat{{\epsilon}}_\theta\left(x_k , c,k\right)=(1+v) {\epsilon}_\theta\left(x_k , c, k\right)-v{\epsilon}_\theta\left(x_k , \varnothing,  k\right),$
% \end{equation}
where $\varnothing$ is the null context and $v$ controls the balance between sample quality and diversity.
% In practice, the unconditioned model is obtained by applying dropout on the context embedding.
Alternatively, in \emph{classifier-guided} diffusion \cite{dhariwal2021diffusion}, a separate classifier \(h_\phi(c \vert \mathbf{x}_k, k)\) is trained, and the gradient \(\nabla_\mathbf{x} \log h_\phi(c \vert \mathbf{x}_k)\) is used for the classifier-guided sampling as follows: $\bar{{\epsilon}}_\theta(\mathbf{x}_k, k) = {\epsilon}_\theta(x_k, k) - \sqrt{1 - \bar{\alpha}_k} \; v \nabla_{\mathbf{x}_k} \log h_\phi(c \vert \mathbf{x}_k).$

\noindent\textbf{Diffusion Planning.}
Offline RL is a setting where an agent aims to learns an optimal policy from a fixed, previously collected dataset without further interaction with the environment \cite{levine2020offline, fu2020d4rl}. This problem can be framed as a sequence modeling task \cite{chen2021decision, janner2021offline}.
Recently, diffusion-based planners \cite{janner2022diffuser, ajay2023is} have utilized DPMs to generate trajectories which can address the challenges of Offline RL, as discussed in \cite{levine2020offline}. 
One typical example of diffusion planning is \emph{Diffuser} \cite{janner2022diffuser}, which utilizes expressive DPMs to model trajectories in the following form:
\begin{equation}
\label{eqn:traj}
{\tau}=\left[\begin{array}{llll}
{s}_0 & {s}_1 & \ldots & {s}_H \\
{a}_0 & {a}_1 & \ldots & {a}_H
\end{array}\right],
\end{equation}
where $H$ is the planning horizon. The model is optimized based on Equation \ref{eqn:ddpm}, with ${\epsilon}_{\theta}(\tau_k, k)$ being modeled by U-Nets \cite{ronneberger2015u}, chosen for their non-autoregressive, temporally local, and equivariant characteristics. 
A separate model $\mathcal{J}_\phi$ is trained to predict the cumulative rewards, and the gradients of $\mathcal{J}_\phi$ are used to guide the trajectory following the classifier-guided sampling procedure.  
Another example is \emph{Decision Diffuser} \cite{ajay2023is}, which adopts a classifier-free approach \cite{Ho2022ClassifierFreeDG} and utilizes reward information as context to generate high-return trajectories. In this work, We adopt the classifier-free approach for its flexibility in incorporating contextual information, a key requirement of our method.

\noindent\textbf{Inversion for Image Manipulation.}
In the domain of \emph{generative adversarial networks} (GANs) \cite{goodfellow2014generative}, manipulating images often involves finding the corresponding latent representation of a given image, a process known as \emph{inversion} \cite{xia2022gan,zhu2016generative}. This can be achieved through optimization-based techniques \cite{abdal2019image2stylegan, abdal2020image2stylegan, gu2020image}, which directly optimize a latent vector to recreate the target image when passed through the GAN, or through the use of encoders \cite{richardson2021encoding, zhu2020domain, tov2021designing}.
Similarly, in the domain of DPMs, inversion enables image manipulations such as cross-image interpolations and semantic editing in \emph{DALL-E 2} \cite{ramesh2022hierarchical}. Lastly, textual inversion \cite{gal2022image} represents visual concepts as novel tokens in a frozen text-to-image model, enabling personalized embeddings.

\noindent\textbf{Preference Learning.}
It has proven to be effective to leverage relative human judgments through pairwise preference labels for optimizing human preferences without direct access to the reward function.
This approach shows significant success in various natural language processing tasks, such as translation \cite{kreutzer2018reliability}, summarization \cite{stiennon2020learning,ziegler2019fine}, story-telling \cite{ziegler2019fine}, and instruction-following \cite{ouyang2022training,ramamurthy2022reinforcement}.
It typically learns a reward function using a preference model like the Bradley-Terry model \cite{bradley1952rank}, and subsequently trains the model using RL algorithms \cite{williams1992simple,schulman2017proximal} to maximize the learned reward.
\emph{Direct policy optimization} (DPO) has been proposed as an alternative to directly align the policy with human preferences and learn from collected data without a separate reward model \cite{rafailov2023direct}. DPO variants \cite{azar2023general, ethayarajh2024kto, xu2024contrastive, zhao2023slic} have shown great alignment with human preferences that matches or surpasses reward-based methods.
In the domain of RL, learning policies from preferences has been studied, as designing a suitable reward function can be challenging.
Various approaches have been proposed \cite{lee2021pebble, knox2022models, hejna2023few, ibarz2018reward, christiano2017deep, kim2023preference} that learn a reward function from trajectory segment pairs.
% Recently, DPO has also been incorporated into this domain \cite{an2023direct}.

%===============================================================================
\begin{figure*}[t]
    \centering
    \includegraphics[width=0.99\textwidth]{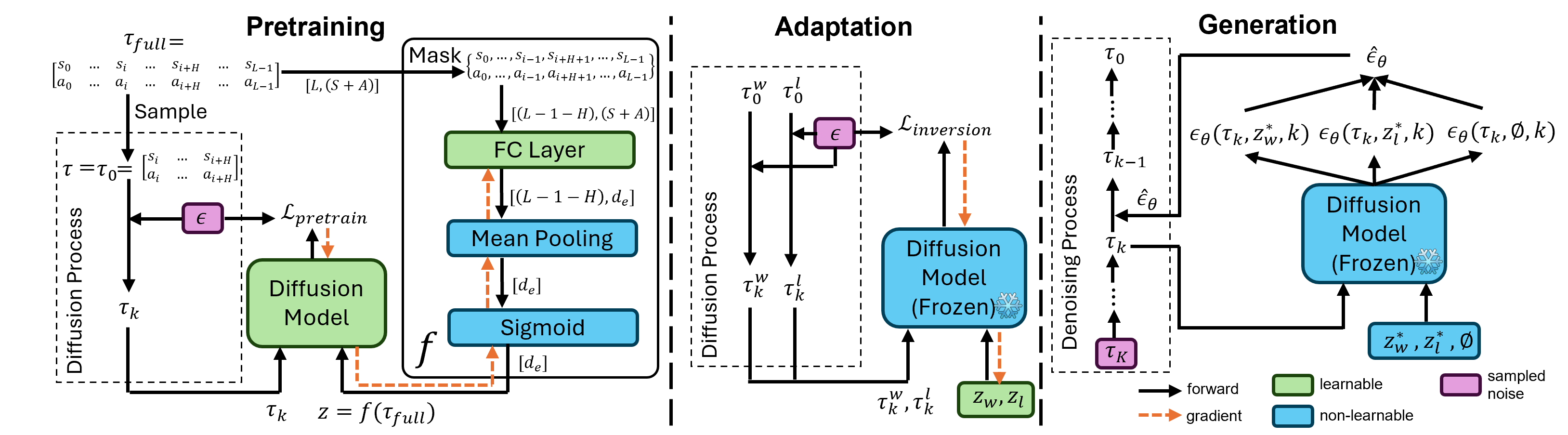}
    \caption{\textbf{Overview of the proposed method.} (Left) Pretraining: A placeholder for preference latent embedding (PLE), $z$, is co-trained with the diffusion model, without reward supervision. (Middle) Adaptation: With diffusion model weights frozen, PLEs are aligned to user labelled query pairs via preference inversion. (Right) Generation: Conditional sampling with learned PLEs generate trajectories that
match the users' preference.}
    \label{fig:arch}
\end{figure*}

\section{Methodology}

To enable rapid adaptation of our pretrained model to individual user preferences, we introduce the concept of \emph{\textbf{preference latent embeddings}} (PLE), denoted by \(z\). 
PLEs are low-dimensional vectors that encode human preferences efficiently. Our method comprises three distinct stages, as illustrated in Figure \ref{fig:arch}: \textbf{pretraining, adaptation, and generation}. 

During \textbf{pretraining}, we train the diffusion model without reward supervision, initializing a placeholder for the PLE to establish a general-purpose generative model.  
In \textbf{adaptation}, a small set of human preference labels fine-tunes the model, identifying the PLE that aligns with user preferences in a low-dimensional representation.  
Finally, in \textbf{generation}, the learned PLE guides trajectory generation to match encoded human preferences. We now detail each stage.

\subsection{Pretraining with Masked Trajectories}

The pretraining stage of our method addresses two concurrent goals: training a general-purpose generative model that comprehensively understands the task domain without reward supervision, and initializing a meaningful representation for the PLE placeholder. To achieve the first goal, we employ the decision diffuser \cite{ajay2023is}, which excels in training on offline trajectory datasets without explicit reward requirements. Its ability to incorporate additional context is crucial for our second goal.

For the second goal, we posit that each preference correlates with a set of similar trajectories. Consequently, we aim to map similar trajectories to similar embeddings to give structure for the PLE placeholder. We propose a learnable mapping function: $f : \mathbb{R}^{L \times (S + A)} \to \mathbb{R}^{d_e}, z := f(\tau_\text{full})$, where $\tau_\text{full}$ represents the full trajectory from which a sub-trajectory $\tau$ is sampled. $L$ denotes the full trajectory length, $S$ and $A$ are the dimensions of the state and action spaces respectively, and $d_e$ is the PLE dimension, a tunable hyperparameter.
The mapping $f$ consists of a sequence of operations as visualized in Figure \ref{fig:arch} (left). Initially, we apply a fixed mask to $\tau_\text{full}$ to prevent information leakage about $\tau$. This is followed by a transformation of state-action features into the latent space using a learnable feed-forward layer. We then apply mean pooling to mitigate time leakage and accommodate variable horizon lengths. Normalization is achieved via sigmoid activation, a choice motivated by the adaptation process described in the subsequent section. Finally, we compose these differentiable components to obtain the PLE placeholder, $z$.
The resulting $z$ is then fed into the decision diffuser as context, training end-to-end to achieve both goals simultaneously with the following objective:
\begin{equation*}
\label{eqn:pretrain}
\resizebox{\linewidth}{!}{
$\mathcal{L}_\text{pretrain}(\theta)=\mathbb{E}_{k \sim[1, K], x_0 \sim q, {\epsilon} \sim \mathcal{N}(0, I)}\left[\left\|{\epsilon}-{\epsilon}_\theta\left(\tau_k,  f(\tau_\text{full}), k \right )\right\|^2\right].$}
\end{equation*}
The proposed objective serves two crucial purposes in our approach: constructing a representation for the PLE placeholder that groups similar trajectories, and pretraining the model to generate trajectories adhering to the offline dataset's distribution. However, the model remains unaligned with specific user preferences at this stage.

\subsection{Adaptation via Preference Inversion}

During adaptation, our goal is to quickly identify the PLE that aligns with the user's preferences. To achieve this, we utilize a small set of human preference labels to partially fine-tune the pretrained model, rather than performing full fine-tuning. This approach is made possible by the placeholder PLE trained during the pretraining step. We begin the adaptation with a randomly initialized, learnable PLE, $z$. The sigmoid activation applied during the pretraining stage enables us to choose a prior bounded between 0 and 1 for this initialization. To align $z$, we freeze all weights of the diffuser model and backpropagate the loss gradients towards $z$. We refer to this PLE alignment process as \textbf{\emph{preference inversion}}, drawing an analogy to the inversion methods for image manipulation described in Section \ref{sec:background}.

To design a loss function that leverages pairwise preference labels, we sub-categorize the PLE into two types: winner PLE \(z_w\), and loser PLE \(z_l\).
This enables us to optimize and obtain learned preferences \(z_w^*\) and \(z_l^*\) based on the reconstruction loss of the respective winner and loser trajectories, $x^w$ and $x^l$ as follows:
\begin{equation*}
\label{eqn:adapt}
\begin{split}
\mathcal{L}_\text{inversion}(z^w, z^l)&=\mathbb{E}_{k \sim[1, K], x_0 \sim q, {\epsilon} \sim \mathcal{N}(0, I)}\left[\left\|{\epsilon}-{\epsilon}_\theta\left(\tau^w_k, z_w, k \right 
 )\right\|^2 \right. \\
 & \left. + \left\|{\epsilon}-{\epsilon}_\theta\left(\tau^l_k,  z_l, k \right 
 )\right\|^2\right].
\end{split}
\end{equation*}

where $\theta$ is fixed. A key advantage of having a frozen pretrained model is that it does not require the additional constraints to remain aligned to the base model, as is the case with RLHF or DPO. During training, it is possible to simultaneously optimize $z_w$ and $z_l$ within a single batch, since their gradients do not interfere with each other.

\subsection{Generating Preferred Trajectories}
To sample a trajectory aligned with human preferences, we utilize a linear combination of the winner and loser PLEs, similar to the approach used in \cite{Ho2022ClassifierFreeDG} to predict noise:
\begin{equation*}
% \label{eq:sample}
\small
\begin{split}
 % &\hat{{\epsilon}}_\theta\left(x_t , z, k\right) |_{z=z_w^*, z_l^*}= 
&\hat{{\epsilon}}_\theta\left(x_t , z_w^*, z_l^*, k \right)  
 =(1+v)\dot{{\epsilon}}_\theta\left(x_t , z_w^*, z_l^*, k\right)- v {\epsilon}_\theta\left(x_t , \varnothing, k\right), \\
 \text{where } & {\dot{{\epsilon}}}_\theta\left(x_t , z_w^*, z_l^*, k\right)=(1+u) {\epsilon}_\theta\left(x_t , z_w^* , k\right)- u {\epsilon}_\theta\left(x_t , z_l^*, k\right).
 \end{split}
\end{equation*}
Here, $v$ and $u$ are hyper-parameters. $v$ controls the strength of the guidance, while $u$ controls the influence of the loser information. To gain an intuition, we rewrite ${\dot{{\epsilon}}}_\theta\left(x_t ,  z_w^*, z_l^*, k\right)= {\epsilon}_\theta\left(x_t , z_w^* , k\right) + u ({\epsilon}_\theta\left(x_t , z_w^* , k\right) - {\epsilon}_\theta\left(x_t , z_l^*, k\right))$, which shows that we are pushing the score estimations away from the loser, originating at the winner.
This allows us to efficiently leverage the pretrained model while quickly adapting to individual user preferences using a small amount of preference data. By learning only the low-dimensional PLE while keeping our pretrained model fixed, we reduce computational cost and enhance the stability of the adaptation process compared to fine-tuning the entire model. The overall proposed method is illustrated in the Figure \ref{fig:arch}.

%===============================================================================

\section{Experiments}

We comprehensively evaluate the effectiveness of our method in integrating user preferences, examining pretraining performance, the impact of preference queries and \( z_l^* \), and adaptation stability. Additionally, we assess its ability to capture human preferences from diverse, high-reward queries using a custom dataset. For a strong benchmark, we compare against diverse baselines.

\begin{itemize}%[leftmargin=*,noitemsep,topsep=0pt,parsep=0pt,partopsep=0pt]

\item \textbf{Diffuser:} A pretrained diffusion-based planner~\cite{Janner2022PlanningWD} representing the distribution of training dataata but not adapted to user labels. 

\item \textbf{Guided Sampling:} 
Following RLHF, we train a reward model using the Bradley-Terry model \cite{bradley1952rank}, but employ classifier-guidance sampling \cite{dhariwal2021diffusion}. %, eliminating the need for further optimization with PPO \cite{schulman2017proximal}. 

\item \textbf{Finetuning (Full):} This baseline directly fine-tunes the pretrained Diffuser model using DPO~\cite{rafailov2023direct} with $\beta=5000$.

\item \textbf{Finetuning (LoRA):} This baseline provides a more direct comparison with our proposed method, where partial fine-tuning is performed. To achieve this, we utilize LoRA \cite{hu2021lora} with a rank of $r=8$. 

\item \textbf{Preference Transformers:} Transformer-based architecture for modeling preferences over trajectories \cite{kim2023preference}.

\item \textbf{{Preference Inversion} (Proposed):} Our method partially fine-tunes the pretrained model to retrieve the user preference context. The pretraining stage utilizes a diffuser conditioned on masked trajectories. 
\end{itemize}

All hyperparameters related to the diffusion model follow those in ~\cite{Janner2022PlanningWD}.

\noindent\textbf{Experimental Setup.} To examine the effectiveness of various personalization methods in automated decision-making systems, we tested our approach on a preference learning benchmark \cite{kim2023preference} that utilizes challenging control tasks from the d4rl dataset \cite{fu2020d4rl} in an offline setting. Specifically, we used the Hopper, HalfCheetah, and Walker2D tasks from the d4rl dataset, focusing on the medium-expert and medium-replay settings. 
For the preference labels, we follow \cite{kim2023preference}, where query pairs (pairs of trajectory segments) are randomly sampled from the D4RL offline dataset. Within each pair, the trajectory segment with the higher return is designated as the winner, with the other segment consequently labeled as the loser.
% we collected random query pairs without access to the task reward and assigned the winner to the query with the higher reward. 
To ensure proper convergence of our large dataset, all baselines underwent 1 million updates for pretraining. During the alignment stage, we fixed \(N_\text{adapt} = 5000\) updates for our main experiments and also conducted ablation studies on \(N_\text{adapt}\). Pretraining utilizes the full dataset for its respective tasks. During evaluation, we sample each method for 100 episodes across 5 different random seeds within their designated environments.

\subsection{Latent Space Analysis}
\begin{figure}[h]
    \centering
    \begin{subfigure}{0.48\linewidth}
        \centering
        \includegraphics[width=\linewidth]{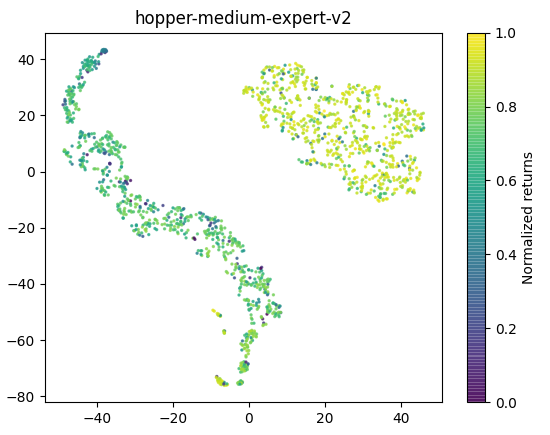}
    \end{subfigure}
    \begin{subfigure}{0.48\linewidth}
        \centering
        \includegraphics[width=\linewidth]{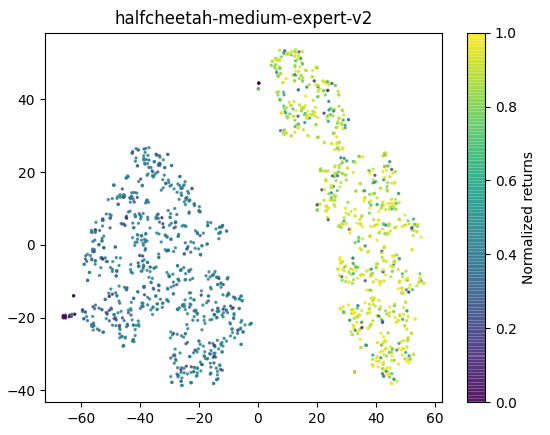}   
    \end{subfigure}    
    \begin{subfigure}{0.48\linewidth}
        \centering
        \includegraphics[width=\linewidth]{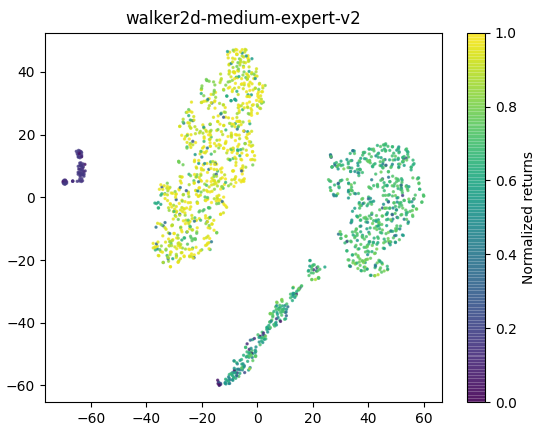}
    \end{subfigure}
    % \par\smallskip    
    \begin{subfigure}{0.48\linewidth}
        \centering
        \includegraphics[width=\linewidth]{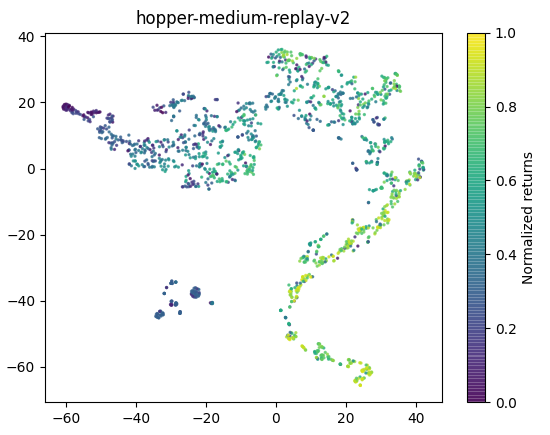}
    \end{subfigure}
    \begin{subfigure}{0.48\linewidth}
        \centering
        \includegraphics[width=\linewidth]{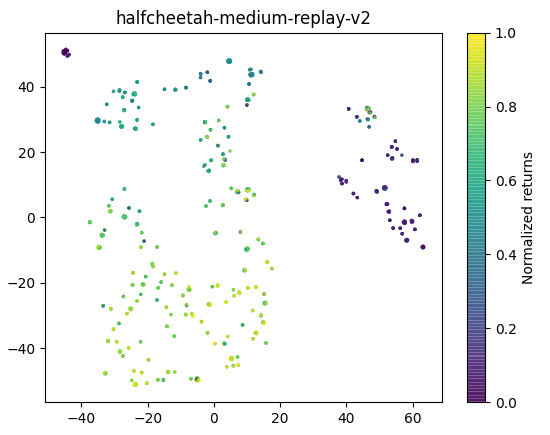}   
    \end{subfigure}    
    \begin{subfigure}{0.48\linewidth}
        \centering
        \includegraphics[width=\linewidth]{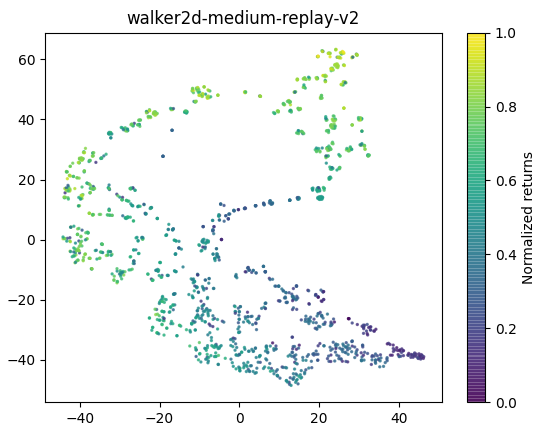}
    \end{subfigure}

\caption{\textbf{Latent space analysis}: We visualize t-SNE plots of PLEs post-pretraining, where each point represents a trajectory, and color intensity reflects its normalized score. The smooth gradient in return distribution indicates that our pretraining effectively structures the PLE space.
}
\label{fig:latent}
% \vspace{-20pt}
\end{figure}

To understand the impact of our proposed mapping in the pretraining for the PLE placeholder, we conducted a visualization of the latent space representation. Using their respective pretrained models, we sampled 1000 random trajectories from each dataset and obtained the PLE, $z$. These embeddings were then projected onto a two-dimensional space using t-SNE (with perplexity set to 30) as seen in Figure \ref{fig:latent}, with color intensity representing the normalized score of the corresponding masked trajectory. 
The normalized score is defined as $\text{normalized score} = 100 \times \frac{\text{score} - \text{random score}}{\text{expert score} - \text{random score}}$ as in \cite{fu2020d4rl}.
Examining the latent space of the medium-expert dataset, we observed distinct clusters representing low, medium, and high returns, respectively. These clusters were well separated and mostly linearly separable. In contrast, the medium-replay dataset, which consists of the entire replay buffer with a continual range of returns, exhibited a different pattern. The latent space reflected this continuous return distribution, showing not distinct clusters but rather a smooth transition of the latent embeddings based on their return values. 

Overall, our proposed pretraining enables the PLE placeholder to be structured in an organized and meaningful manner, demonstrating that our proposed mapping, $f$, is able to organize similar trajectories close together.
This organized latent space could accelerate the preference inversion process by first navigating the loss landscape to a local region of similar trajectories, and then refining the search for a more precise alignment with the true reward (human preference).

\begin{figure}[t]
    \centering
    \includegraphics[width=0.99\linewidth]{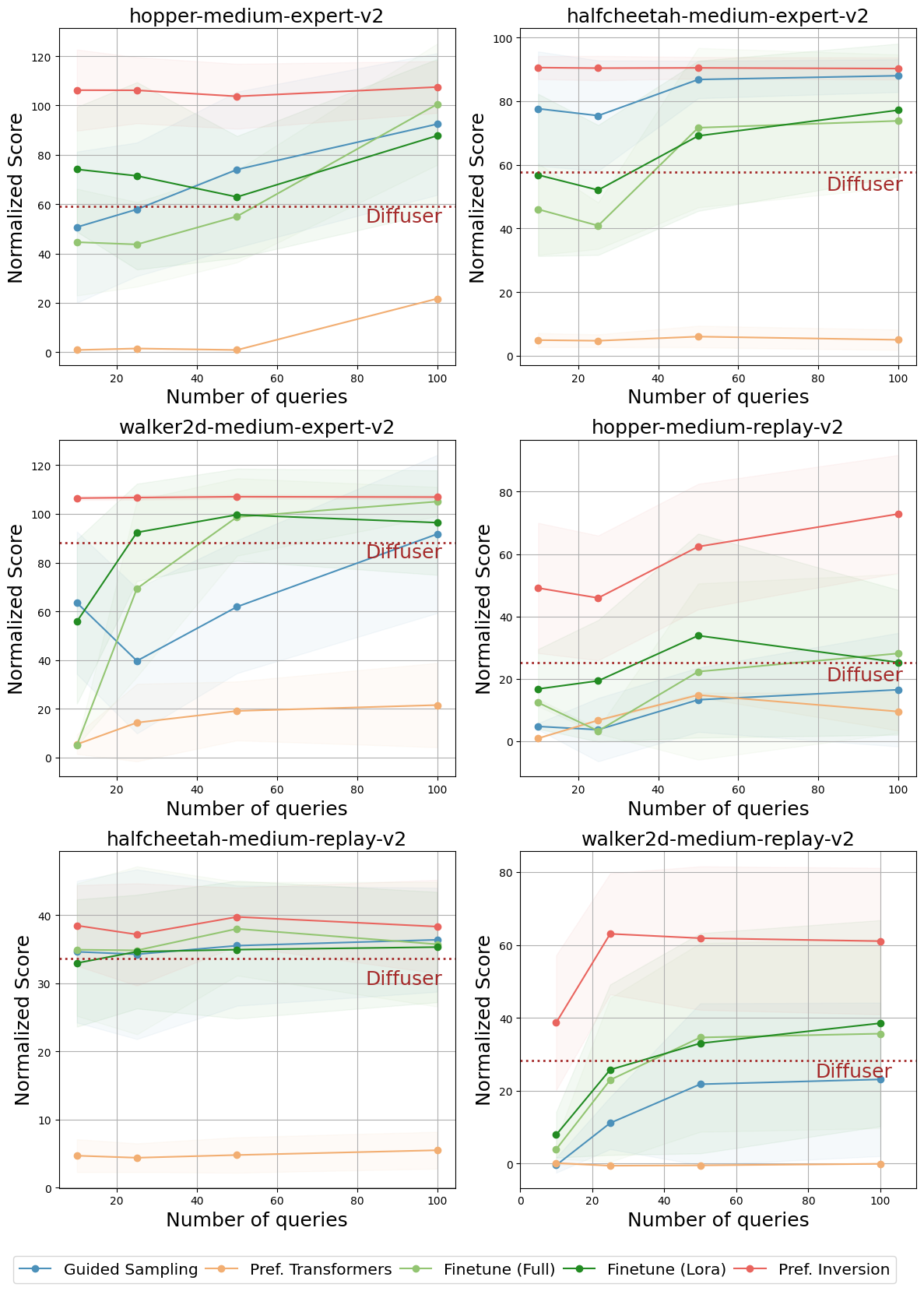}
    \caption{\textbf{Main results evaluated over different numbers of queries across six control tasks report the normalized score.}}
    \label{fig:main}
% \vspace{-10pt}
\end{figure}

\label{sec:result}
\subsection{Main Results}
We compare our method against the baselines and assess the impact of the number of query pairs, $N_\text{query}$. We evaluate each model with $N_\text{query}$ values of 10, 25, 50, and 100, analyzing their ability to align with user preferences using a small set of human-annotated queries and a limited number of updates ($N_\text{adapt}=5000$).
For the main experiment, we set $d_e=16$ and $u=0.02$.

Our experimental results (Figure \ref{fig:main}) show that our method consistently outperforms baselines, with a growing advantage as queries decrease. This is particularly evident in hopper-medium-replay and walker2d-medium-replay tasks. No baseline consistently outperforms others: LoRA fine-tuning closely matches full fine-tuning, with slight gains at lower queries (N=10, 25) and mixed results for N=50, 100, while Preference Transformers perform worst, likely due to training from scratch without pretrained models. Baselines (except Preference Transformers) surpass the pretrained Diffuser when \( N_{\text{query}} > 50 \) but often fail with fewer queries, leading to negative adaptation. In contrast, our method consistently outperforms Diffuser, maintaining high performance even at \( N_{\text{query}} = 10 \). Performance correlates positively with \( N_{\text{query}} \), with 50 queries generally sufficient across tasks, except for hopper-medium-expert, which requires 100. These results validate our resource-efficient approach, reducing labeled data needs with minimal updates.

\begin{figure*}[t]
    % \centering
    \begin{minipage}{0.69\textwidth}
    \begin{subfigure}[t]{0.32\linewidth}
        \centering
        \includegraphics[width=\linewidth]{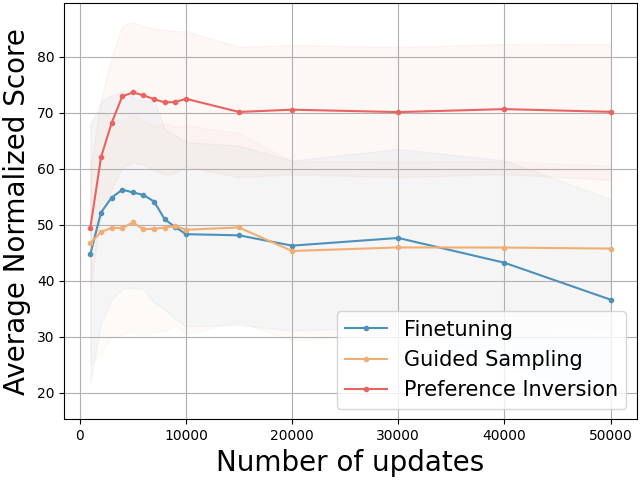}
        \caption{Adaptation stability across \(N_\text{adapt}\).}
        \label{fig:ab1}
    \end{subfigure}
    % \hspace{10pt}
    \begin{subfigure}[t]{0.32\linewidth}
        \centering
        \includegraphics[width=\linewidth]{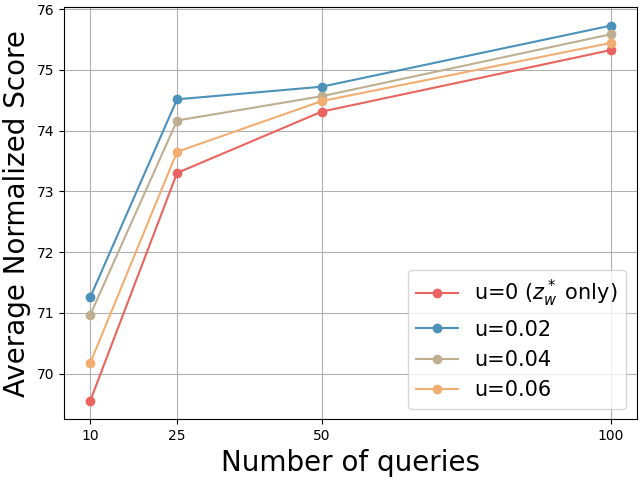}
        \caption{The impact of utilizing loser PLE, $z_l^*$ for sampling}
        \label{fig:ab2}
    \end{subfigure}
    % \hspace{10pt}
    \begin{subfigure}[t]{0.32\linewidth}
        \centering
        \includegraphics[width=\linewidth]{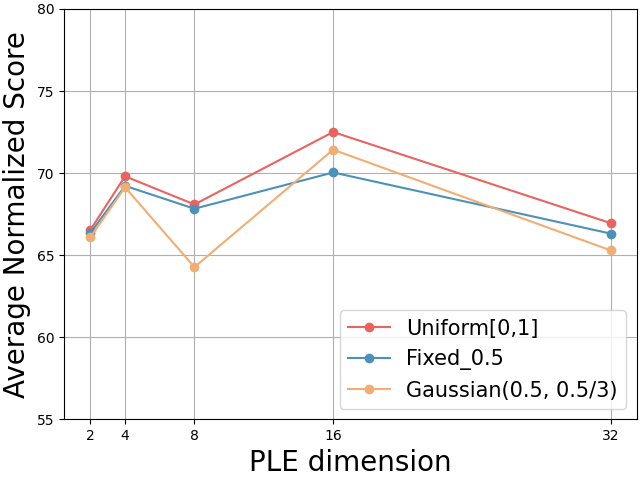}
        \caption{Choice of different priors for initialization, choice of PLE dimension}
        \label{fig:ab3}
    \end{subfigure}
    \caption{\textbf{A series of ablation experiments.} The average normalized score is reported across all tasks and $N_\text{query}$, except for the loser PLE analysis, where averaging is performed across tasks only.
}
    \label{fig:ab}
    \end{minipage}
    \hfill \hfill \hfill
% \end{figure*}
% \begin{figure*}[htpb]
    \begin{minipage}{0.28\textwidth}
    % \vspace{-5pt}
    % \centering
    \includegraphics[width=0.99\linewidth]{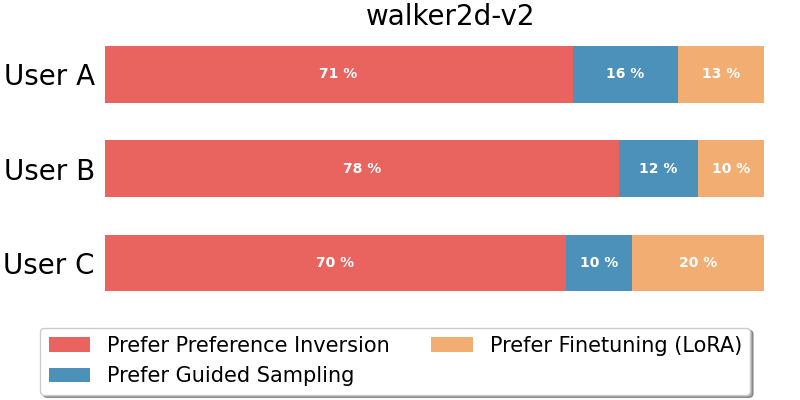}
    % \vspace{-5pt}
    \caption{\textbf{Survey Results on Real Human Preference.} Users select their preferred trajectories from samples generated by top 2 baselines and our proposed method.
    }
    \label{fig:human}
    % \vspace{-20pt}
    \end{minipage}
\end{figure*}

\begin{figure*}[htpb]
% \vspace{-20pt}
    \begin{subfigure}[t]{0.3\linewidth}
        \centering
        \includegraphics[width=\linewidth]{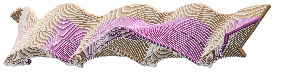}
        \caption{User A: `Back leg swing up high and body learning forward'}
        \label{fig:human2}
    \end{subfigure}
    \hfill
    \begin{subfigure}[t]{0.3\linewidth}
        \centering
        \includegraphics[width=\linewidth]{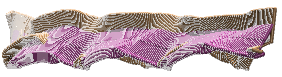}
        \caption{User B: `Body upright, front knee bent, rear leg 45degrees'}
        \label{fig:human1}
    \end{subfigure}
    \hfill
    \begin{subfigure}[t]{0.3\linewidth}
        \centering
        \includegraphics[width=\linewidth]{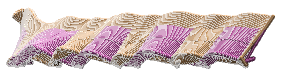}
        \caption{User C: `Gentle hopping with moderately fast strides'}
        \label{fig:human3}
    \end{subfigure}
    \caption{\textbf{Trajectories generated using our proposed method, an aligned model conditioned on user's respective PLE.} The generated samples closely match each user's description of their preference.} 
    \label{fig:human_description}
    % \vspace{-20pt}
\end{figure*}

\subsection{Ablation Studies}
We perform a series of ablation experiments to gain deeper insights into the relative importance of different design choices and determine the sensitivity of our approach to variations in model components and hyperparameters. 

\noindent\textbf{Number of adaptation steps.}
Figure \ref{fig:ab1} shows that all methods peak around $N_\text{adapt}=5000$. Preference inversion and guided sampling remain stable after minor drops at 15000 and 20000 updates, respectively. However, the performance of finetuning consistently declines after peaking, with rapid deterioration after $N_\text{adapt}=30000$, suggesting excessive deviation from the base model. The stability of our method is advantageous for practical applications, where the optimal stopping point is often unknown in advance.

\noindent\textbf{Loser PLE.}
Figure \ref{fig:ab2} shows that incorporating the loser PLE, $z_l^*$ with $u>0$, consistently improves the sampling performance compared to using only the winner PLE with $u=0$. The improvements peak at $u=0.02$, resulting in $1\sim3\%$ gains across various $N_\text{query}$ values, and gradually diminish as $u$ decreases further. Utilizing $z_l^*$ provides a small boost when $N_\text{query}$ is high, but notably  enhances the sampling when $N_\text{query}$ is low.

\noindent\textbf{Choice of prior and PLE dimension.}
Given that our PLE $z$ is constrained to the range [0, 1], we test three different priors for initialising $z$ within this interval: $\text{Uniform}[0,1]$, $\text{Gaussian}(0.5, 0.5/3)$, and $\text{Fixed}\_{0.5}$. Figure \ref{fig:ab3} demonstrates that all three settings perform comparably well, with the uniform prior slightly outperforming the others. Similarly, varying the PLE dimension \(d_e\) across 2, 4, 8, 16, and 32 consistently yields good results, demonstrating relative insensitivity to this hyper-parameter, with a slight advantage observed at \(d_e=16\).

\subsection{Real Human Preference on Quality Diversity Dataset}
Previous experiments, following the design of \cite{kim2023preference}, focus on recovering a hidden task reward. While useful for validating human preferences, it \textbf{prioritizes high-reward over diversity} and may not fully capture practical scenarios. In contrast, practical decision-making often involves selection from a \textbf{diverse set of high-reward} trajectories. To better reflect this, we design a new experiment based on Quality Diversity (QD) \cite{pugh2016quality,conti2018improving, wu2022quality}, which in policy learning refers to an algorithm's ability to discover diverse, high-performing policies with distinct behaviors.
To implement this, we train a set of QD policies based on \cite{wu2022quality}, generating a diverse dataset of 750 high-reward Walker2D episodes for model pretraining without preference alignment. Additionally, we use QD policies to create query pairs and gather preference labels from three users. They are instructed to maintain a consistent selection strategy and provide written descriptions of their decision criteria.
We then adapt the pretrained policy to each user's preference labels, consisting of 100 query pairs each.

For evaluation, we generate 100 trajectories per baseline and ask users to choose the closest match to their preference criteria. The survey results, shown in Figure \ref{fig:human}, indicate that our proposed method receives the vast majority of votes, demonstrating its effectiveness in capturing human preferences. Figure~\ref{fig:human_description} qualitatively displays the sampled trajectories from the aligned model, generated using preference inversion, which closely match the users' descriptions. 
% We share the details of our new dataset in Appendix \ref{apx:dataset} and hope it can be used for future benchmarks. 
This real human preference dataset provides a good initial indication of our method's practical real-world applicability. To establish more robust findings, we plan to survey additional users in our future work.

%===============================================================================

\section{Conclusion}
\label{sec:conclusion}
This work presents a novel approach that enables a policy to quickly adapt to a small human preference dataset. It consists of pretraining followed by adaptation on latent embeddings via preference inversion for rapid alignment. Evaluation results demonstrate that our method adapts more accurately to human preferences with minimal preference labels, outperforming baselines in both offline datasets and our custom dataset with real human labels. This promising method shows potential for further applications across diverse settings.

%%%%%%%%%%%%%%%%%%%%%%%%%%%%%%%%%%%%%%%%%%%%%%%%%%%%%%%%%%%%%%%%%%%%%%%%%%%%%%%%

%%%%%%%%%%%%%%%%%%%%%%%%%%%%%%%%%%%%%%%%%%%%%%%%%%%%%%%%%%%%%%%%%%%%%%%%%%%%%%%%


\section*{ACKNOWLEDGMENT}
This study is supported under RIE2020 Industry Alignment Fund – Industry Collaboration Projects (IAF-ICP) Funding Initiative, as well as cash and in-kind contribution from the industry partner(s).

\bibliographystyle{IEEEtran}
\bibliography{IEEEexample}
% \end{thebibliography}

\end{document}